\documentclass{article}

    \PassOptionsToPackage{numbers}{natbib}

\usepackage[final]{neurips_2022}

\usepackage[utf8]{inputenc} %
\usepackage[T1]{fontenc}    %
\usepackage{hyperref}       %
\usepackage{url}            %
\usepackage{booktabs}       %
\usepackage{amsfonts}       %
\usepackage{nicefrac}       %
\usepackage{microtype}      %
\usepackage{xcolor}         %

\usepackage{graphicx}
\usepackage{mathtools}
\usepackage{amsmath}
\usepackage{amsthm}
\usepackage{amssymb}
\usepackage{commath}
\usepackage{subfigure}
\usepackage{wrapfig}

\newcommand{\xo}{\mathbf{x}_o}
\newcommand{\xu}{\mathbf{x}_u}
\newcommand{\z}{\mathbf{z}}
\newcommand{\x}{\mathbf{x}}

\DeclareMathOperator*{\argmax}{argmax}
\DeclareMathOperator*{\argmin}{argmin}

\newtheorem{theorem}{Theorem}[section]

\title{Posterior Matching for Arbitrary Conditioning}

\author{%
  Ryan~R.~Strauss \\
  Department of Computer Science\\
  UNC at Chapel Hill\\
  \texttt{rrs@cs.unc.edu} \\
  \And
  Junier~B.~Oliva \\
  Department of Computer Science\\
  UNC at Chapel Hill\\
  \texttt{joliva@cs.unc.edu} \\
}

\begin{document}

\maketitle

\begin{abstract}
Arbitrary conditioning is an important problem in unsupervised learning, where we seek to model the conditional densities $p(\xu \mid \xo)$ that underly some data, for all possible non-intersecting subsets \mbox{$o, u \subset \{1, \dots , d\}$}. However, the vast majority of density estimation only focuses on modeling the joint distribution $p(\x)$, in which important conditional dependencies between features are opaque. We propose a simple and general framework, coined Posterior Matching, that enables Variational Autoencoders (VAEs) to perform arbitrary conditioning, without modification to the VAE itself. Posterior Matching applies to the numerous existing VAE-based approaches to joint density estimation, thereby circumventing the specialized models required by previous approaches to arbitrary conditioning. We find that Posterior Matching is comparable or superior to current state-of-the-art methods for a variety of tasks with an assortment of VAEs (e.g.~discrete, hierarchical, VaDE).
\end{abstract}

\section{Introduction}

Variational Autoencoders (VAEs) \cite{kingma2013auto} are a widely adopted class of generative model that have been successfully employed in numerous areas \cite{bowman2015generating,gregor2015draw,maaloe2016auxiliary,razavi2019generating,han2019variational}. Much of their appeal stems from their ability to probabilistically represent complex data in terms of lower-dimensional latent codes.

Like most other generative models, VAEs are typically designed to model the joint data distribution, which communicates likelihoods for particular configurations of all features at once. This can be useful for some tasks, such as generating images, but the joint distribution is limited by its inability to explicitly convey the conditional dependencies between features. In many cases, conditional distributions, which provide the likelihood of an event given some known information, are more relevant and useful. Conditionals can be obtained in theory by marginalizing the joint distribution, but in practice, this is generally not analytically available and is expensive to approximate. 

Easily assessing the conditional distribution over \textit{any subset} of features is important for tasks where decisions and predictions must be made over a varied set of possible information.
For example, some medical applications may require reasoning over: the distribution of \textit{blood pressure} given \textit{age} and \textit{weight}; or the distribution of \textit{heart-rate} and \textit{blood-oxygen level} given \textit{age}, \textit{blood pressure}, and \textit{BMI}; etc.
For flexibility and scalability, it is desirable for a \textit{single} model to provide all such conditionals at inference time.
More formally, this task is known as \textit{arbitrary conditioning}, where the goal is to model the conditional density $p(\xu \mid \xo)$ for any arbitrary subsets of unobserved features $\xu$ and observed features $\xo$. In this work, we show, by way of a simple and general framework, that traditional VAEs can perform arbitrary conditioning, without modification to the VAE model itself.

Our approach, which we call Posterior Matching, is to model the distribution $p(\z \mid \xo)$ that is induced by some VAE, where $\z$ is the latent code. In other words, we consider the distribution of latent codes given partially observed features. We do this by having a neural network output an approximate partially observed posterior $q(\z \mid \xo)$. In order to train this network, we develop a straightforward maximum likelihood estimation objective and show that it is equivalent to maximizing $p(\xu \mid \xo)$, the quantity of interest. Unlike prior works that use VAEs for arbitrary conditioning, we do not make special assumptions or optimize custom variational lower bounds. \textit{Rather, training via Posterior Matching is simple, highly flexible, and without limiting assumptions on approximate posteriors (e.g.,~$q(\z \mid \xo)$ need not be reparameterized and can thus be highly expressive).}

We conduct several experiments in which we apply Posterior Matching to various types of VAEs for a myriad of different tasks, including image inpainting, tabular arbitrary conditional density estimation, partially observed clustering, and active feature acquisition. We find that Posterior Matching leads to improvements over prior VAE-based methods across the range of tasks we consider.

\section{Background}

\paragraph{Arbitrary Conditioning}
A core problem in unsupervised learning is \textit{density estimation}, where we are given a dataset $\mathcal{D} = \{ \x^{(i)} \}_{i=1}^N$ of \textit{i.i.d.}~samples drawn from an unknown distribution $p(\x)$ and wish to learn a model that best approximates the probability density function $p$. A limitation of only learning the joint distribution $p(\x)$ is that it does not provide direct access to the conditional dependencies between features. \textit{Arbitrary conditional density estimation} \cite{ivanov2018variational,li2020acflow,strauss2021arbitrary} is a more general task where we want to estimate the conditional density $p(\xu \mid \xo)$ for all possible subsets of observed features \mbox{$o \subset \{1, \dots, d\}$} and unobserved features \mbox{$u \subset \{1, \dots , d\}$} such that $o$ and $u$ do not intersect. Here, $\xo \in \mathbb{R}^{|o|}$ and $\xu \in \mathbb{R}^{|u|}$. Estimation of joint or marginal likelihoods is a special case where $o = \emptyset$. Note that, while not strictly necessary for arbitrary conditioning methods \cite{li2020acflow,strauss2021arbitrary}, we assume $\mathcal{D}$ is fully observed, a requirement for training traditional VAEs.

\paragraph{Variational Autoencoders}

Variational Autoencoders (VAEs) \cite{kingma2013auto} are a class of generative models that assume a generative process in which data likelihoods are represented as \mbox{$p(\x) = \int p(\x \mid \z) p(\z) \dif \z$}, where $\z$ is a latent variable that typically has lower dimensionality than the data $\x$. A tractable distribution that affords easy sampling and likelihood evaluation, such as a standard Gaussian, is usually imposed on the prior $p(\z)$. These models are learned by maximizing the \textit{evidence lower bound} (ELBO) of the data likelihood:
\begin{equation*}
    \log p(\x) \geq \mathbb{E}_{\z \sim q_\psi(\cdot \mid \x)} [ \log p_\phi(\x \mid \z) ] - \text{KL}(q_\psi(\z \mid \x) \mid \mid p(\z)),
\end{equation*}
where $q_\psi(\z \mid \x)$ and $p_\phi(\x \mid \z)$ are the encoder (or approximate posterior) and decoder of the VAE, respectively. The encoder and decoder are generally neural networks that output tractable distributions (e.g.,~a multivariate Gaussian). In order to properly optimize the ELBO, samples drawn from $q_\psi(\z \mid \x)$ must be differentiable with respect to the parameters of the encoder (often called the \textit{reparameterization trick}). After training, a new data point $\hat{\x}$ can be easily generated by first sampling $\z$ from the prior, then sampling $\hat{\x} \sim p_\phi(\cdot \mid \z)$.

\section{Posterior Matching}
\begin{wrapfigure}{r}{0.5\linewidth}
    \vspace{-0.8cm}
    \centering
    \includegraphics[trim={0mm 5mm 0mm 2mm},clip,width=\linewidth]{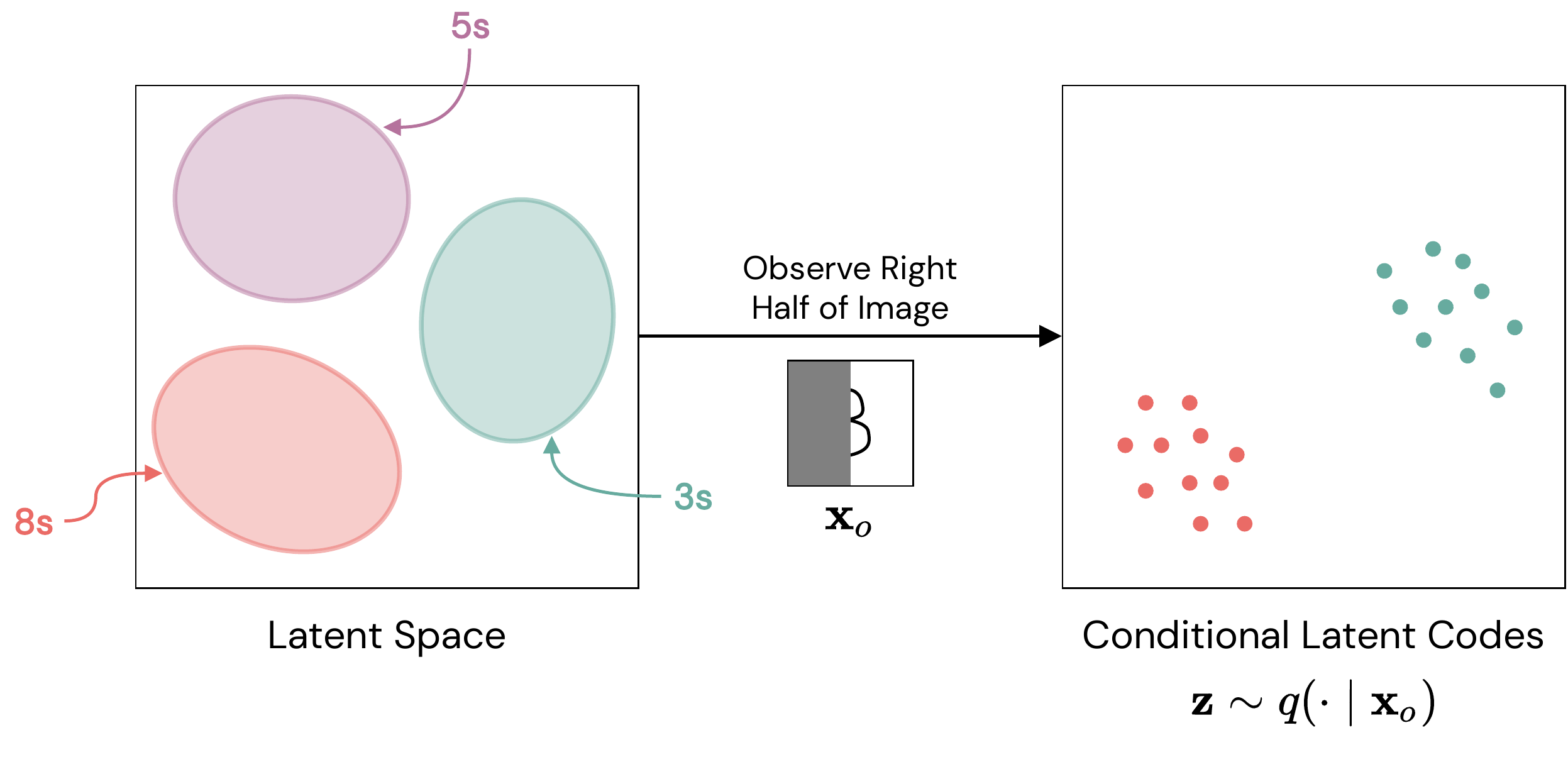}
    \caption{A two-dimensional VAE latent space (left) that represents the distribution of codes of handwritten 3s, 5s, and 8s. Conditioning on the subset of pixels shown in the center should then result in a distribution over latent codes that only correspond to 3s and 8s, as shown on the right.}
    \label{fig:latent-example}
\end{wrapfigure}In this section we describe our framework, coined Posterior Matching, to model the underlying arbitrary conditionals in a VAE. In many respects, Posterior Matching \textit{cuts the Gordian knot} to uncover the conditional dependencies. Following our insights, we show that our approach is \textit{direct} and \textit{intuitive}. Notwithstanding, we are the first to apply this direct methodology for arbitrary conditionals in VAEs and are the first to connect our proposed loss with arbitrary conditional likelihoods $p(\xu \mid \xo)$. Note that we are \textit{not} proposing a new type of VAE. Rather, we are formalizing a simple and intuitive methodology that can be applied to numerous existing (or future) VAEs.

\subsection{Motivation}

Let us begin with a motivating example, depicted in \autoref{fig:latent-example}. Suppose we have trained a VAE on images of handwritten 3s, 5s, and 8s. This VAE has thus learned to represent these images in a low-dimensional latent space. Any given code (vector) in this latent space represents a distribution over images in the original data space, which can be retrieved by passing that code through the VAE's decoder. Some regions in the latent space will contain codes that represent 3s, some will represent 5s, and some will represent 8s. There is typically only an interest in mapping from a given image $\x$ to a distribution over the latent codes that could represent that image, i.e.,~the posterior $q(\z \mid \x)$. 
\textit{However, we can just as easily ask which latent codes are feasible having only observed part of an image.}

For example, if we only see the right half the image shown in \autoref{fig:latent-example}, we know the digit could be a 3 or an 8, but certainly not a 5. Thus, the distribution over latent codes that could correspond to the full image, that is $p_\psi(\z \mid \xo)$ (where $\psi$ is the encoder's parameters), should only include regions that represent 3s or 8s. Decoding any sample from $p_\psi(\z \mid \xo)$ will produce an image of a 3 or an 8 that aligns with what has been observed.

The important \emph{insight} is that we can think about how conditioning on $\xo$ changes the distribution over latent codes without explicitly worrying about what the (potentially higher-dimensional and more complicated) conditional distribution over $\xu$ looks like. Once we know $p_\psi(\z \mid \xo)$, we can easily move back to the original data space using the decoder.

\subsection{Approximating the Partially Observed Posterior}

The partially observed approximate posterior of interest is not readily available, as it is implicitly defined by the VAE:
\begin{equation} \label{eq:true-pop}
    p_\psi(\z \mid \xo) = \mathbb{E}_{\xu \sim p(\cdot \mid \xo)} \Big[ q_\psi(\z \mid \xo, \xu) \Big],
\end{equation}
where $q_\psi(\z \mid \xo, \xu) = q_\psi(\z \mid \x)$ is the VAE's encoder. Thus, we introduce a neural network in order to approximate it.

\begin{figure}[t]
    \centering
    \includegraphics[width=0.74\linewidth]{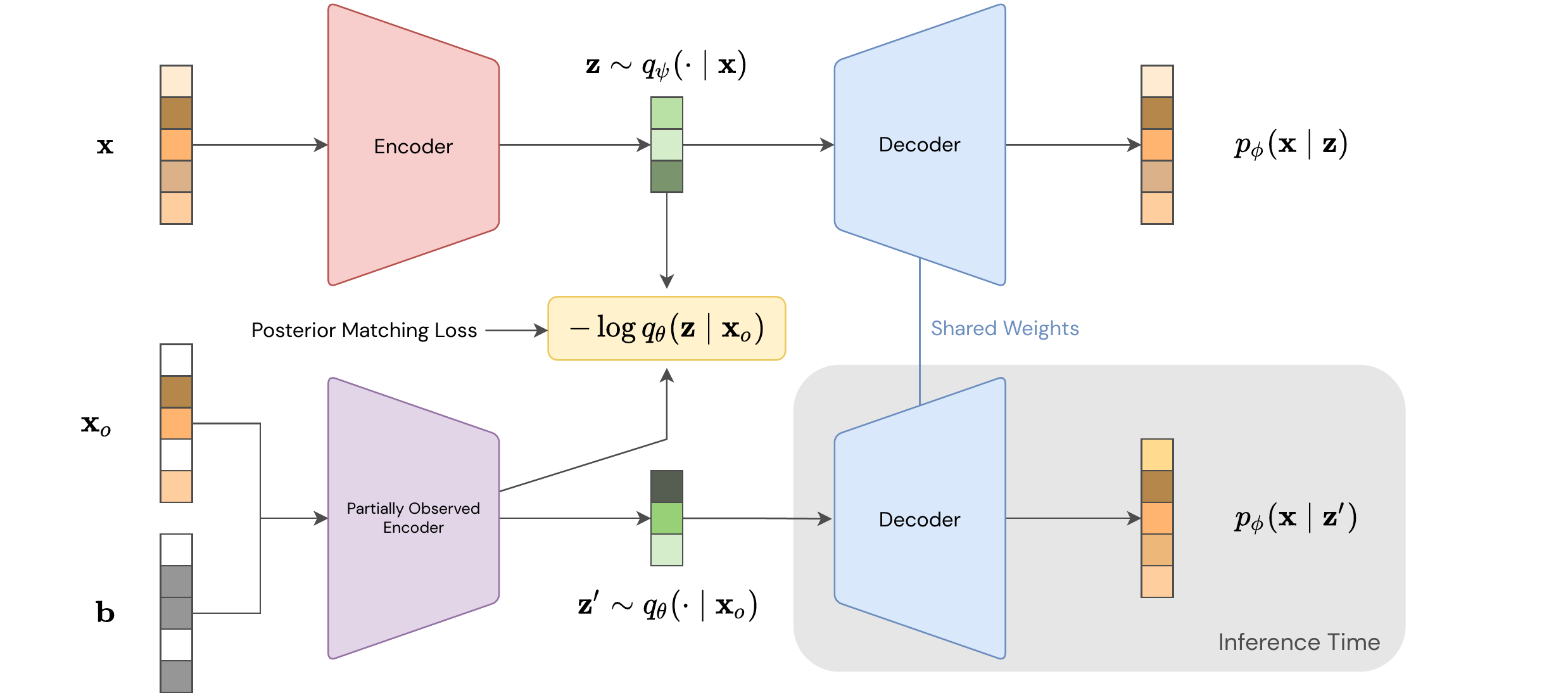}
    \caption{Overview of the Posterior Matching framework. The partially observed encoder can be appended to any existing VAE to enable arbitrary conditioning. At inference time, we can decode samples from $q_\theta(\z \mid \xo)$ to perform imputation or compute $p(\xu \mid \xo)$.}
    \label{fig:model}
\end{figure}

Given a network that outputs the distribution $q_{\theta}(\z \mid \xo)$ (i.e.~the partially observed encoder in \autoref{fig:model}), we now discuss our approach to training it. Our approach is guided by the priorities of simplicity and generality. We minimize (with respect to $\theta$) the following likelihoods, where the samples are coming from our target distribution as defined in \autoref{eq:true-pop}:
\begin{equation} \label{eq:objective-ll}
    \mathbb{E}_{\xu \sim p(\cdot \mid \xo)} \Big[ \mathbb{E}_{\z \sim q_\psi(\cdot \mid \xo, \xu)} [ - \log q_\theta(\z \mid \xo) ] \Big].
\end{equation}
We discuss how this is optimized in practice in \autoref{sec:implementation}.

Due to the relationship between negative log-likelihood minimization and KL-divergence minimization \cite{bishop2006pattern},
we can interpret \autoref{eq:objective-ll} as %
minimizing: 
\begin{equation} \label{eq:objective-kl}
    \mathbb{E}_{\xu \sim p(\cdot \mid \xo)} \Big[ \text{KL}\left(q_\psi(\z \mid \xo, \xu) \mid \mid q_\theta(\z \mid \xo)\right) \Big].
\end{equation}

We can directly minimize the KL-divergence in \autoref{eq:objective-kl} if it is analytically available between the two posteriors, for instance if both posteriors are Gaussians. However, \autoref{eq:objective-ll} is more general in that it allows us to use more expressive (e.g.,~autoregressive) distributions for $q_\theta(\z \mid \xo)$ with which the KL-divergence cannot be directly computed. This is important given that $p_\psi(\z \mid \xo)$ is likely to be complex (e.g.,~multimodal) and not easily captured by a Gaussian (as in \autoref{fig:latent-example}). Importantly, there is no requirement for $q_\theta(\z \mid \xo)$ to be reparameterized, which would further limit the class of distributions that can be used. There is a high degree of flexibility in the choice of distribution for the partially observed posterior. Note that this objective does not utilize the decoder.

\subsection{Connection with Arbitrary Conditioning}

While the Posterior Matching objective from \autoref{eq:objective-ll} and \autoref{eq:objective-kl} is intuitive, it is not immediately clear how this approach relates back to the arbitrary conditioning objective of maximizing \mbox{$p(\xu \mid \xo)$}. We formalize this connection in \autoref{thm:pm-ac} (see Appendix for proof).

\newpage

\begin{theorem} \label{thm:pm-ac}
Let $q_{\psi}(\z \mid \x)$ and $p_{\phi}(\x \mid \z)$ be the encoder and decoder, respectively, for some VAE. Additionally, let $q_{\theta}(\z \mid \xo)$ be an approximate partially observed posterior. Then minimizing
$\mathbb{E}_{\xu \sim p(\cdot \mid \xo)} \left[ \text{KL} \left( q_{\psi}(\z \mid \xo, \xu) \mid \mid q_{\theta}(\z \mid \xo) \right) \right]$
is equivalent to minimizing
\begin{equation} \label{eq:alt-obj}
    \mathbb{E}_{\xu \sim p(\cdot \mid \xo)} \Big[ - \log p_{\theta, \phi}(\xu \mid \xo) + \text{KL} \left( q_{\psi}(\z \mid \xo, \xu) \mid \mid q_{\theta}(\z \mid \xo, \xu) \right) \Big],
\end{equation}
with respect to the parameters $\theta$.
\end{theorem}

The first term inside the expectation in \autoref{eq:alt-obj} gives us the explicit connection back to the arbitrary conditioning likelihood $p(\xu \mid \xo)$, which is being maximized when minimizing \autoref{eq:alt-obj}. The second term acts as a sort of regularizer by trying to make the partially observed posterior match the VAE posterior when conditioned on all of $\x$ --- intuitively, this makes sense as a desirable outcome.

\subsection{Implementation} \label{sec:implementation}

A practical training loss follows quickly from \autoref{eq:objective-ll}. For the outer expectation, we do not have access to the true distribution $p(\xu \mid \xo)$, but for a given instance $\x$ that has been partitioned into $\xo$ and $\xu$, we do have one sample from this distribution, namely $\xu$. So we approximate this expectation using $\xu$ as a single sample. This type of single-sample approximation is common with VAEs, e.g., when estimating the ELBO. For the inner expectation, we have access to $q_\psi(\z \mid \x)$, which can easily be sampled in order to estimate the expectation. In practice, we generally use a single sample for this as well. This gives us the following Posterior Matching loss:
\begin{equation}
    \mathcal{L}_{\text{PM}}(\x, o, \theta, \psi) = - \mathbb{E}_{\z \sim q_\psi(\cdot \mid \x)} \left[\log q_\theta(\z \mid \xo) \right],
\end{equation}
where $o$ is the set of observed feature indices. During training, $o$ can be randomly sampled from a problem-specific distribution for each minibatch.

\autoref{fig:model} provides a visual overview of our approach. In practice, we represent $\xo$ as a concatenation of $\x$ that has had unobserved features set to zero and a bitmask $\mathbf{b}$ that indicates which features are observed. This representation has been successful in other arbitrary conditioning models \cite{li2020acflow,strauss2021arbitrary}. However, this choice is not particularly important to Posterior Matching itself, and alternative representations, such as set embeddings, are valid as well. 

As required by VAEs, samples from $q_\psi(\z \mid \x)$ will be reparameterized, which means that minimizing $\mathcal{L}_{\text{PM}}$ will influence the parameters of the VAE's encoder in addition to the partially observed posterior network. In some cases, this may be advantageous, as the encoder can be guided towards learning a latent representation that is more conducive to arbitrary conditioning. However, it might also be desirable to train the VAE independently of the partially observed posterior, in which case we can choose to stop gradients on the samples $\z \sim q_\psi(\cdot \mid \x)$ when computing $\mathcal{L}_{\text{PM}}$.

Similarly, the partially observed posterior can be trained against an existing pretrained VAE. In this case, the parameters of the VAE's encoder and decoder are frozen, and we only optimize $\mathcal{L}_{\text{PM}}$ with respect to $\theta$. Otherwise, we jointly optimize the VAE's ELBO and $\mathcal{L}_{\text{PM}}$.

We emphasize that there is a high degree of flexibility with the choice of VAE, i.e.~we have not imposed any unusual constraints. However, there are some potentially limiting practical considerations that have not been explicitly mentioned yet. First, the training data must be fully observed, as with traditional VAEs, since $\mathcal{L}_{\text{PM}}$ requires sampling $q_\psi(\z \mid \x)$. However, given that the base VAE requires fully observed training data anyway, this is generally not a relevant limitation for our purposes. Second, it is \textit{convenient} in practice for the VAE's decoder to be factorized, i.e.~$p(\x \mid \z) = \prod_i p(x_i \mid \z)$, as this allows us to easily sample from $p(\xu \mid \z)$ (sampling $\xu$ is less straightforward with other types of decoders). However, it is standard practice to use factorized decoders with VAEs, so this is ordinarily not a concern. We also note that, while useful for easy sampling, a factorized decoder is \textit{not necessary} for optimizing the Posterior Matching loss, which does not incorporate the decoder.

\subsection{Posterior Matching Beyond Arbitrary Conditioning} \label{sec:beyond-ac}

\begin{wrapfigure}{R}{0.5\linewidth}
    \centering
    \vspace{-0.55cm}
    \includegraphics[trim={0mm 0mm 0mm 0mm},clip,width=\linewidth]{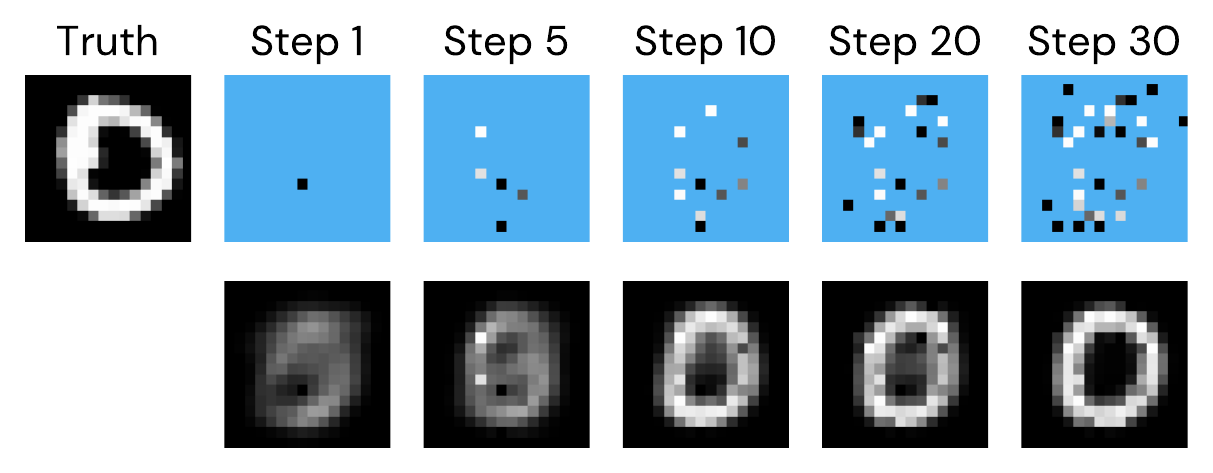}
    \caption{Example acquisitions from our CNN model using the lookahead posteriors (see \autoref{sec:afa-exp}). Top row shows $\xo$, and blue pixels are unobserved. Bottom row shows imputation of $\xu$.}
    \label{fig:acquisition}
    \vspace{-0.5cm}
\end{wrapfigure}

The concept of matching VAE posteriors is quite general and has other uses beyond the application of arbitrary conditioning. We consider one such example, which still has ties to arbitrary conditioning, in order to give a flavor for other potential uses. 

A common application of arbitrary conditioning is \textit{active feature acquisition} \cite{gong2019icebreaker,li2021active,ma2018eddi}, where informative features are sequentially acquired on an instance-by-instance basis. In the unsupervised case, the aim is to acquire as few features as possible while maximizing the ability to reconstruct the remaining unobserved features (see \autoref{fig:acquisition} for example).

One approach to active feature acquisition is to greedily select the feature that will maximize the expected amount of information to be gained about the currently unobserved features \cite{li2021active,ma2018eddi}. For VAEs, \citet{ma2018eddi} show that this is equivalent to selecting each feature according to
\begin{equation} \label{eq:info-gain-z}
    \argmax_{i \in u} H(\z \mid \xo) - \mathbb{E}_{x_i \sim p(\cdot \mid \xo)} \Big[ H(\z \mid \xo, x_i) \Big] =
    \argmin_{i \in u}  \mathbb{E}_{x_i \sim p(\cdot \mid \xo)} \Big[ H(\z \mid \xo, x_i) \Big].
\end{equation}
For certain families of posteriors, such as multivariate Gaussians, the entropies in \autoref{eq:info-gain-z} can be analytically computed. 
In practice, approximating the expectation in \autoref{eq:info-gain-z} is done via entropies of the posteriors
$
    p^{(i)}(\z \mid \xo) \equiv \mathbb{E}_{x_i \sim p_{\theta,\phi}(\cdot \mid \xo)} \Big[ q_\theta(\z \mid \xo, x_i) \Big],
$
where samples from $p_{\theta,\phi}(x_i \mid \xo)$ are produced by first sampling $\z \sim q_\theta(\cdot \mid \xo)$ and then passing $\z$ through the VAE's decoder $p_\phi(x_i \mid \z)$ (we call $p^{(i)}(\z \mid \xo)$ the ``lookahead'' posterior for feature $i$, since it is obtained by imagining what the posterior will look like after one acquisition into the future).
Hence, computing the resulting entropies requires one network evaluation per sample of $x_i$ to encode $z$, for $i\in u$. Thus, if using $k$ samples for each $x_i$, 
each greedy step will be $\Omega(k \cdot |u|)$, which may be prohibitive in high dimensions.

In analogous fashion to the Posterior Matching approach that has already been discussed, we can train a neural network to directly output the lookahead posteriors for all features at once. The Posterior Matching loss in this case is
\begin{equation} \label{eq:lookahead-loss}
    \mathcal{L}_{\text{PM-Lookahead}}(\x, o, u, \omega, \theta, \phi) = \sum_{i \in u} \mathbb{E}_{x_i \sim p_{\theta, \phi}(\cdot \mid \xo)} \Bigg[ \mathbb{E}_{\z \sim q_{\theta}(\cdot \mid \xo, x_i)} \Big[ -\log q_\omega^{(i)}(\z \mid \xo) \Big] \Bigg],
\end{equation}
where $\omega$ is the parameters of the lookahead posterior network. 
In practice, we train a single shared network with a final output layer that outputs the parameters of all $q_\omega^{(i)}(\z \mid \xo)$.
Note that given the distributions $q_\omega^{(i)}(\z \mid \xo)$ for all $i$, computing the greedy acquisition choice consists of doing a forward evaluation of our network, then choosing the feature $i \in u$ such that the entropy of $q_\omega^{(i)}(\z \mid \xo)$ is minimized. In other words, we may bypass the individual samples of $x_i$, and use a single shared network for a faster acquisition step. In this setting, we let $q_\omega^{(i)}(\z \mid \xo)$ be a multivariate Gaussian so that the entropy computation is trivial. See Appendix for a diagram of the entire process. This use of Posterior Matching leads to large improvements in the computational efficiency of greedy active feature acquisition (demonstrated empirically in \autoref{sec:afa-exp}).

\section{Prior Work}

A variety of approaches to arbitrary conditioning have been previously proposed. ACE is an autoregressive, energy-based method that is the current state-of-the-art for arbitrary conditional likelihood estimation and imputation, although it can be computationally intensive for very high dimensional data \cite{strauss2021arbitrary}. ACFlow is a variant of normalizing flows that can give analytical arbitrary conditional likelihoods \cite{li2020acflow}. Several other methods, including Sum-Product Networks \cite{poon2011sum,butz2019deep}, Neural Conditioner \cite{belghazi2019learning}, and Universal Marginalizer \cite{douglas2017universal}, also have the ability to estimate conditional likelihoods.

\citet{rezende2014stochastic} were among the first to suggest that VAEs can be used for imputation. More recently, VAEAC was proposed as a VAE variant designed for arbitrary conditioning \cite{ivanov2018variational}.
Unlike Posterior Matching, VAEAC is not a general framework and cannot be used with typical pretrained VAEs.
EDDI is a VAE-based approach to active feature acquisition and relies on arbitrary conditioning \cite{ma2018eddi}. The authors introduce a ``Partial VAE'' in order to perform the arbitrary conditioning, which, similarly to Posterior Matching, tries to model $p(\z \mid \xo)$. Unlike Posterior Matching, they do this by maximizing a variational lower bound on $p(\xo)$ using a partial inference network $q(\z \mid \xo)$ (there is no standard VAE posterior $q(\z \mid \x)$ in EDDI). \citet{gong2019icebreaker} use a similar approach that is based on the Partial VAE of EDDI. The major drawback of these methods is that, unlike with Posterior Matching, $q(\z \mid \xo)$ must be reparameterizable in order to optimize the lower bound (the authors use a diagonal Gaussian). Thus, certain more expressive distributions (e.g.,~autoregressive) cannot be used. Additionally, these methods cannot be applied to existing VAEs.
The methods of \citet{ipsen2020not} and \citet{collier2020vaes} are also similar to EDDI, where the former optimizes an approximation of $p(\mathbf{x}_o, \mathbf{b})$ and the latter optimizes a lower bound on $p(\mathbf{x}_o \mid \mathbf{b})$. \citet{ipsen2020not} also focuses on imputation for data that is missing ``not at random'', a setting that is outside the focus of our work.

There are also several works that have considered learning to identify desirable regions in latent spaces. \citet{engel2017latent} start from a pretrained VAE, but then train a separate GAN \cite{goodfellow2014generative} with special regularizers to do their conditioning. They only condition on binary vectors, $\mathbf{y}$, that correspond to a small number of predefined attributes, whereas we allow for conditioning on arbitrary subsets of continuous features $\xo$ (a more complicated conditioning space). Also, their resulting GAN does not make the likelihood $q(\z \mid \mathbf{y})$ available, whereas Posterior Matching directly  (and flexibly) models $q(\z \mid \xo)$, which may be useful for downstream tasks (e.g.~\autoref{sec:afa-exp}) and likelihood evaluation (see Appendix). Furthermore, Posterior Matching trains directly through KL, without requiring an additional critic. \citet{whang2021composing} learn conditional distributions, but not \textit{arbitrary} conditional distributions (a much harder problem). They also consider normalizing flow models, which are limited to invertible architectures with tractable Jacobian determinants and latent spaces that have the same dimensionality as the data (unlike VAEs). \citet{cannella2020projected} similarly do conditional sampling from a model of the joint distribution, but are also restricted to normalizing flow architectures and require a more expensive MCMC procedure for sampling.

\section{Experiments}

In order to empirically test Posterior Matching, we apply it to a variety of VAEs aimed at different tasks. We find that our models are able to match or surpass the performance of previous specialized VAE methods. All experiments were conducted using JAX \cite{jax2018github} and the DeepMind JAX Ecosystem \cite{deepmind2020jax}. Code is available at \url{https://github.com/lupalab/posterior-matching}.

Our results are dependent on the choice of VAE, and the particular VAEs used in our experiments were \emph{not} the product of extensive comparisons and did not undergo thorough hyperparameter tuning --- that is not the focus of this work. With more carefully selected or tuned VAEs, and as new VAEs continue to be developed, we can expect Posterior Matching's downstream performance to improve accordingly on any given task. We emphasize that our experiments span a \emph{diverse set of task, domains, and types of VAE}, wherein Posterior Matching was effective.

\subsection{MNIST}

\begin{figure}[t]
    \centering
    \includegraphics[width=\linewidth]{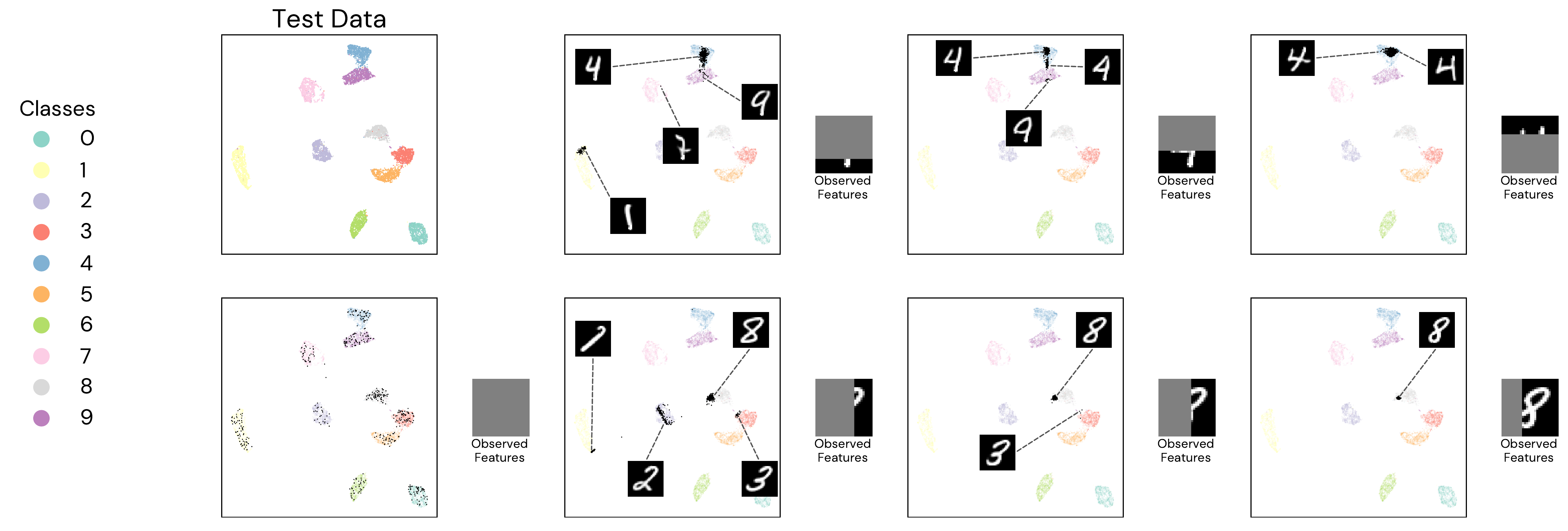}
    \caption{UMAP \cite{mcinnes2018umap} visualization of the latent space of a VAE trained on MNIST. Black dots represent samples from the distribution $q_\theta(\z \mid \xo)$ learned via Posterior Matching. Images of $\xo$ are to the right of their respective plot. Samples from each mode are decoded and shown.}
    \label{fig:mnist-umap}
\end{figure}

In this first experiment, our goal is to demonstrate that Posterior Matching replicates the intuition depicted in \autoref{fig:latent-example}. We do this by training a convolutional VAE with Posterior Matching on the MNIST dataset. The latent space of this VAE is then mapped to two dimensions with UMAP \cite{mcinnes2018umap} and visualized in \autoref{fig:mnist-umap}. In the figure, black points represent samples from $q_\theta(\z \mid \xo)$, and for select samples, the corresponding reconstruction is shown. The encoded test data is shown, colored by true class label, to highlight which regions correspond to which digits. We see that the experimental results nicely replicate our earlier intuitions --- the learned distribution $q_\theta(\z \mid \xo)$ puts probability mass only in parts of the latent space that correspond to plausible digits based on what is observed and successfully captures multimodal distributions (see the second column in \autoref{fig:mnist-umap}).

\subsection{Image Inpainting}

One practical application of arbitrary conditioning is image inpainting, where only part of an image is observed and we want to fill in the missing pixels with visually coherent imputations. As with prior works \cite{ivanov2018variational,li2020acflow}, we assume pixels are missing completely at random.  We test Posterior Matching as an approach to this task by pairing it with both discrete and hierarchical VAEs.

\paragraph{Vector Quantized-VAEs}
We first consider VQ-VAE \cite{oord2017neural}, a type of VAE that is known to work well with images. VQ-VAE differs from the typical VAE with its use of a discrete latent space. That is, each latent code is a grid of discrete indices rather than a vector of continuous values. Because the latent space is discrete, \citet{oord2017neural} model the prior distribution with a PixelCNN \cite{oord2016conditional,salimans2017pixelcnn++} after training the VQ-VAE. We similarly use a conditional PixelCNN to model $q_\theta(\z \mid \xo)$. First, a convolutional network maps $\xo$ to a vector, and that vector is then used as a conditioning input to the PixelCNN. More architecture and training details can be found in the Appendix. We train VQ-VAEs with Posterior Matching for the MNIST, \textsc{Omniglot}, and \textsc{CelebA} datasets. \autoref{tab:inpainting} reports peak signal-to-noise ratio (PSNR) and precision/recall \cite{sajjadi2018assessing} for inpaintings produced by our model. We find that Posterior Matching with VQ-VAE consistently achieves better precision/recall scores than previous models while having comparable PSNR.

\begin{table}[t]
\caption{Peak signal-to-noise ratio (PSNR) and precision/recall scores \cite{sajjadi2018assessing} for image inpaintings. We report mean and standard deviation across five evaluations with different random masks. VAEAC and ACFlow results are taken from \citet{li2020acflow}. Higher is better for all metrics.}
\label{tab:inpainting}
\resizebox{\textwidth}{!}{%
\begin{tabular}{@{}lccccccccc@{}}
\toprule
\multicolumn{1}{c}{} & \multicolumn{3}{c}{MNIST}      & \multicolumn{3}{c}{\textsc{Omniglot}}   & \multicolumn{3}{c}{\textsc{CelebA}}     \\ \cmidrule(l){2-10} 
\multicolumn{1}{c}{} &
  \multicolumn{1}{c}{PSNR} &
  \multicolumn{1}{c}{Precision} &
  \multicolumn{1}{c}{Recall} &
  \multicolumn{1}{c}{PSNR} &
  \multicolumn{1}{c}{Precision} &
  \multicolumn{1}{c}{Recall} &
  \multicolumn{1}{c}{PSNR} &
  \multicolumn{1}{c}{Precision} &
  \multicolumn{1}{c}{Recall} \\ \midrule
VDVAE + PM (ours)  & \textbf{21.603 $\pm$ 0.022} & \textbf{0.996} & \textbf{0.996} & 18.256 $\pm$ 0.038 & \textbf{0.995} & \textbf{0.994} & \textbf{27.190 $\pm$ 0.049} & \textbf{0.995} & \textbf{0.995} \\
VQ-VAE + PM (ours)  & 19.981 $\pm$ 0.021 & 0.989 & 0.993 & 17.954 $\pm$ 0.046 & 0.979 & 0.973 & 25.531 $\pm$ 0.036 & 0.982 & 0.984 \\ \midrule
VAEAC                & 19.613 $\pm$ 0.042 & 0.877 & 0.975 & 17.693 $\pm$ 0.023 & 0.525 & 0.926 & 23.656 $\pm$ 0.027 & 0.966 & 0.967 \\ \midrule
ACFlow               & 17.349 $\pm$ 0.018 & 0.945 & 0.984 & 15.572 $\pm$ 0.031 & 0.962 & 0.971 & 22.393 $\pm$ 0.040 & 0.970 & 0.988 \\
ACFlow+BG            & 20.828 $\pm$ 0.031 & 0.947 & 0.983 & \textbf{18.838 $\pm$ 0.009} & 0.967 & 0.970 & 25.723 $\pm$ 0.020 & 0.964 & 0.987 \\ \bottomrule
\end{tabular}%
}
\end{table}

\paragraph{Hierarchical VAEs}

Hierarchical VAEs \cite{kingma2016improved,sonderby2016ladder,vahdat2020nvae} are a powerful extension of traditional VAEs that allow for more expressive priors and posteriors by partitioning the latent variables into subsets \mbox{$\z = \{\z_1 , \dots , \z_L \}$}. A hierarchy is then created by fractorizing the prior $p(\z) = \prod_i p(\z_i \mid \z_{<i})$ and posterior $q(\z \mid \x) = \prod_i q(\z_i \mid \z_{<i}, \x)$. These models have demonstrated impressive performance on images and can even outperform autoregressive models \cite{child2020very}. Posterior Matching can be naturally applied to hierarchical VAEs, where the partially observed posterior is represented as \mbox{$q(\z \mid \xo) = \prod_i q(\z_i \mid \z_{<i}, \xo)$}.
We adopt the Very Deep VAE (VDVAE) architecture used by \citet{child2020very} and extend it to include the partially observed posterior (see Appendix for training and architecture details). We note that due to our hardware constraints, we trained smaller models and for fewer iterations than \citet{child2020very}. Inpainting results for our VDVAE models are given in \autoref{tab:inpainting}. We see that they achieve better precision/recall scores than the VQ-VAE models and, unlike VQ-VAE, are able to attain better PSNR than ACFlow for MNIST and \textsc{CelebA}. \autoref{fig:inpaintings} shows some example inpaintings, and additional samples are provided in the Appendix. The fact that we see better downstream performance when using VDVAE than when using VQ-VAE is illustrative of Posterior Matching's ability to admit easy performance gains by simply switching to a more powerful base VAE.

\begin{figure}[t]
    \centering
    \valign{#\cr
    \hbox{%
    \subfigure[\textsc{Omniglot}]{\includegraphics[height=1.85cm]{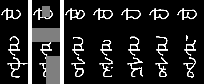}}
    ~\hfill
    \subfigure[\textsc{MNIST}]{\includegraphics[height=1.85cm]{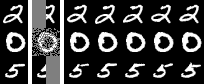}}
    ~\hfill
    \subfigure[\textsc{CelebA}]{\includegraphics[height=1.85cm]{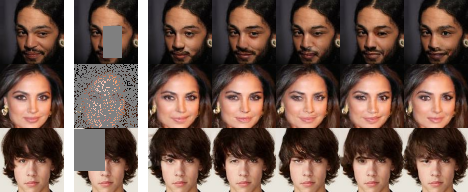}}
    } \cr
    }
    \caption{Example image inpaintings from VDVAE with Posterior Matching. First two columns on the left are $\x$ and $\xo$. Inpainting samples are on the right.}
    \label{fig:inpaintings}
\end{figure}

\subsection{Real-valued Datasets}

We evaluate Posterior Matching on real-valued tabular data, specifically the benchmark UCI repository datasets from \citet{papamakarios2017masked}. We follow the experimental setup used by \citet{li2020acflow} and \citet{strauss2021arbitrary}. In these experiments, we train basic VAE models while simultaneously learning the partially observed posterior. Given the flexibility that Posterior Matching affords, we use an autoregressive distribution for $q_\theta(\z \mid \xo)$. Further details can be found in the Appendix.

\autoref{tab:uci-results} reports the arbitrary conditional log-likelihoods and normalized root-mean-square error (NRMSE) of imputations for our models (with features missing completely at random). Likelihoods are computed using an importance sampling estimate (see Appendix for details). We primarily compare to VAEAC as a baseline in the VAE family, however we also provide results for ACE and ACFlow for reference. We see that Posterior Matching is able to consistently produce more accurate imputations and higher likelihoods than VAEAC. While our models don't match the likelihoods achieved by ACE and ACFlow, Posterior Matching is comparable to them for imputation NRMSE.

\begin{table}[t]
    \centering
    \caption{Test normalized root-mean-square error (NRMSE) and arbitrary conditional log-likelihoods (LL) for UCI datasets. Lower is better for NRMSE, and higher is better for LL. Results for methods other than Posterior Matching are reproduced from \citet{strauss2021arbitrary}. Mean and standard deviation are reported over 5 random observed masks for each instance.}
    \label{tab:uci-results}
    \resizebox{\textwidth}{!}{%
    \begin{tabular}{@{}lcccccccccc@{}}
\toprule
                   & \multicolumn{2}{c}{\textsc{Power}}              & \multicolumn{2}{c}{\textsc{Gas}}               & \multicolumn{2}{c}{\textsc{Hepmass}}             & \multicolumn{2}{c}{\textsc{Miniboone}}          & \multicolumn{2}{c}{BSDS}               \\ \cmidrule(l){2-11} 
                   & NRMSE             & LL                 & NRMSE             & LL                & NRMSE             & LL                  & NRMSE             & LL                 & NRMSE             & LL                 \\ \midrule
Posterior Matching & 0.834 $\pm$ 0.001 & 0.246 $\pm$ 0.002  & 0.330 $\pm$ 0.013 & 5.964 $\pm$ 0.005 & 0.857 $\pm$ 0.000 & -8.963 $\pm$ 0.007  & 0.450 $\pm$ 0.002 & -3.116 $\pm$ 0.175 & 0.573 $\pm$ 0.000 & 77.488 $\pm$ 0.012 \\
VAEAC              & 0.880 $\pm$ 0.001 & -0.042 $\pm$ 0.002 & 0.574 $\pm$ 0.033 & 2.418 $\pm$ 0.006 & 0.896 $\pm$ 0.001 & -10.082 $\pm$ 0.010 & 0.462 $\pm$ 0.002 & -3.452 $\pm$ 0.067 & 0.615 $\pm$ 0.000 & 74.850 $\pm$ 0.005 \\ \midrule
ACE                & 0.828 $\pm$ 0.002 & 0.631 $\pm$ 0.002  & 0.335 $\pm$ 0.027 & 9.643 $\pm$ 0.005 & 0.830 $\pm$ 0.001 & -3.859 $\pm$ 0.005  & 0.432 $\pm$ 0.003 & 0.310 $\pm$ 0.054  & 0.525 $\pm$ 0.000 & 86.701 $\pm$ 0.008 \\
ACE Proposal       & 0.828 $\pm$ 0.002 & 0.583 $\pm$ 0.003  & 0.312 $\pm$ 0.033 & 9.484 $\pm$ 0.005 & 0.832 $\pm$ 0.001 & -4.417 $\pm$ 0.005  & 0.436 $\pm$ 0.004 & -0.241 $\pm$ 0.056 & 0.535 $\pm$ 0.000 & 85.228 $\pm$ 0.009 \\
ACFlow             & 0.877 $\pm$ 0.001 & 0.561 $\pm$ 0.003  & 0.567 $\pm$ 0.050 & 8.086 $\pm$ 0.010 & 0.909 $\pm$ 0.000 & -8.197 $\pm$ 0.008  & 0.478 $\pm$ 0.004 & -0.972 $\pm$ 0.022 & 0.603 $\pm$ 0.000 & 81.827 $\pm$ 0.007 \\
ACFlow+BG          & 0.833 $\pm$ 0.002 & 0.528 $\pm$ 0.003  & 0.369 $\pm$ 0.016 & 7.593 $\pm$ 0.011 & 0.861 $\pm$ 0.001 & -6.833 $\pm$ 0.006  & 0.442 $\pm$ 0.001 & -1.098 $\pm$ 0.032 & 0.572 $\pm$ 0.000 & 81.399 $\pm$ 0.008 \\ \bottomrule
\end{tabular}%
}
\end{table}

\subsection{Partially Observed Clustering}

Probabilistic clustering often views cluster assignments as a latent variable. Thus, when applying Posterior Matching in this setting, we may perform ``partially observed'' clustering, which clusters instances based on a subset of observed features. We consider VaDE, which uses a mixture of Gaussians as the prior, allowing it to do unsupervised clustering by treating each Gaussian component as one of the clusters \cite{jiang2016variational}. Despite differences in how VaDE is trained compared to a classic VAE, training a partially observed encoder via Posterior Matching remains exactly the same.

We train models on both MNIST and \textsc{Fashion MNIST} (see Appendix for experimental details). \autoref{fig:clustering} shows the clustering accuracy of these models as the percentage of (randomly selected) observed features changes. As a baseline, we train a supervised model where the labels are the cluster predictions from the VaDE model when all of the features are observed. We see that Posterior Matching is able to match the performance of the baseline, and even slightly outperform it for low percentages of observed features. Unlike the supervised approach, Posterior Matching has the advantage of being generative.

\begin{figure}[t]
    \centering
    \includegraphics[width=0.49\linewidth]{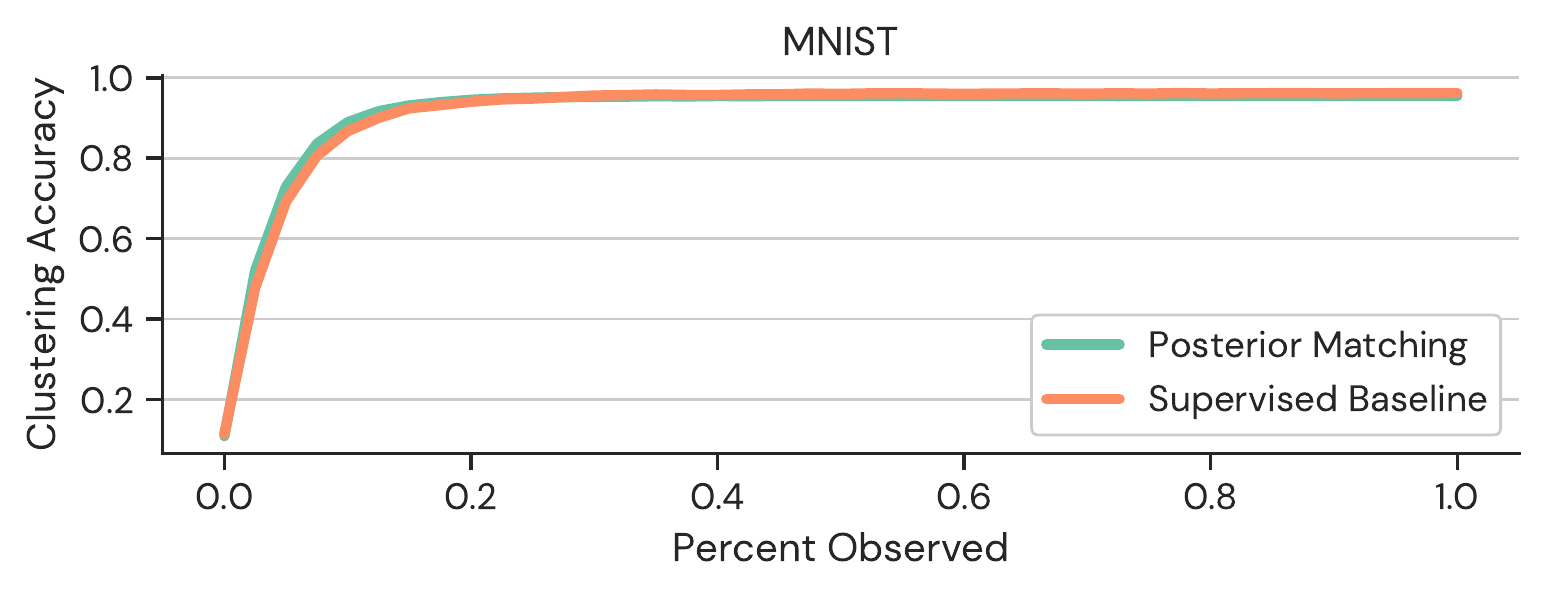}
    \hfill
    \includegraphics[width=0.49\linewidth]{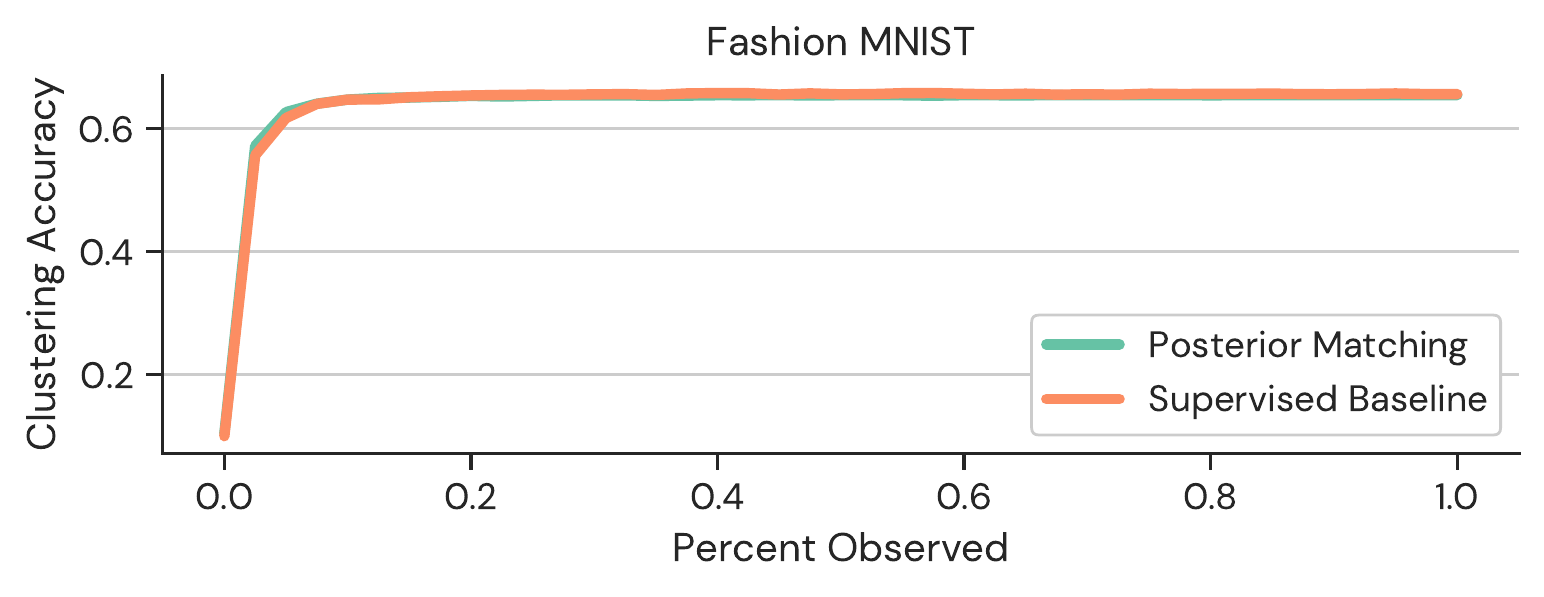}
    \caption{Partially observed clustering accuracy achieved by Posterior Matching with VaDE.}
    \label{fig:clustering}
\end{figure}

\subsection{Very Fast Greedy Feature Acquisition} \label{sec:afa-exp}

As discussed in \autoref{sec:beyond-ac}, we can use Posterior Matching outside of the specific task of arbitrary conditioning. Here, we consider the problem of greedy active feature acquisition. We train a VAE with a Posterior Matching network that outputs the lookahead posteriors described in \autoref{sec:beyond-ac}, using the loss in \autoref{eq:lookahead-loss}. Note that we are also still using Posterior Matching in order to learn $q_\theta(\z \mid \xo)$ and therefore to produce reconstructions. Training details can be found in the Appendix.

\begin{wrapfigure}{R}{0.5\linewidth}
    \centering
    \includegraphics[trim={0mm 0mm 0mm 0mm},clip,width=\linewidth]{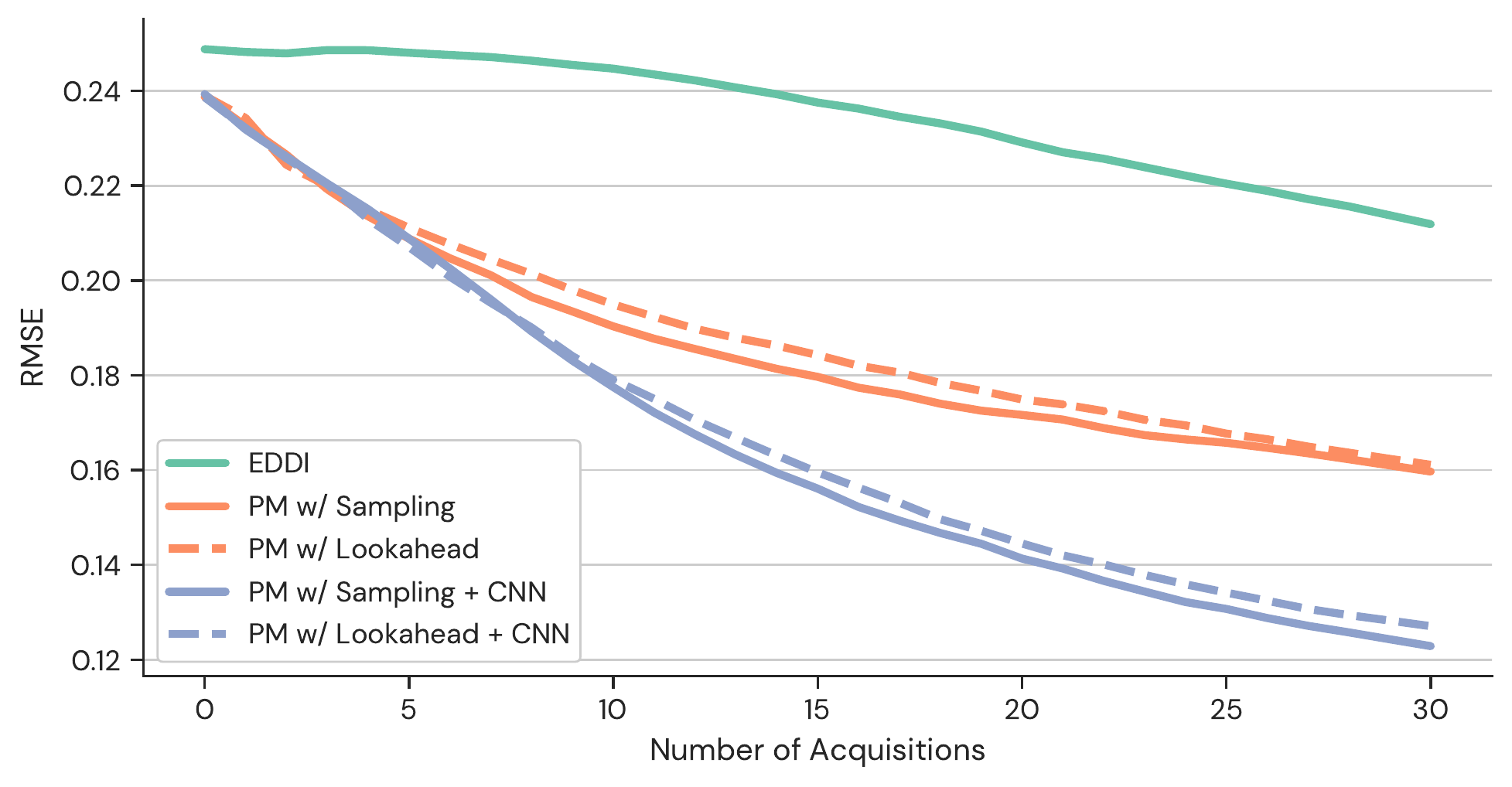}
    \caption{Average RMSE of reconstructions for greedy active feature acquisition.}
    \label{fig:air-rmse}
\end{wrapfigure}

We consider the MNIST dataset and compare to EDDI as a baseline, using the authors' publicly available code. We downscale images to $16 \times 16$ since EDDI has difficulty scaling to high-dimensional data. We also only evaluate on the first 1000 instances of the MNIST test set, as the EDDI code was very slow when computing the greedy acquisition policy. EDDI also uses a particular architecture that is not compatible with convolutions. Thus we train a MLP-based VAE on flattened images in order to make a fair comparison. However, since Posterior Matching does not place any limitations on the type of VAE being used, we also train a convolutional version. For our models, we greedily select the feature to acquire using the more expensive sampling-based approach (similar to EDDI) as well as with the lookahead posteriors (which requires no sampling). In both cases, imputations are computed with an expectation over 50 latent codes, as is done for EDDI. An example acquisition trajectory is shown in \autoref{fig:acquisition}.

\autoref{fig:air-rmse} presents the root-mean-square error, averaged across the test instances, when imputing $\xu$ with different numbers of acquired features. We see that our models are able to achieve lower error than EDDI. We also see that acquiring based on the lookahead posteriors incurs only a minimal increase in error compared to the sampling-based method, despite being far more efficient. Computing the greedy choice with our model using the sampling-based approach takes \mbox{68 ms $\pm$ 917 $\mu$s} (for a single acquisition on CPU). Using the lookahead posteriors, the time is only \mbox{310 $\mu$s $\pm$ 15.3 $\mu$s}, \textbf{a roughly 219x speedup}.

\section{Conclusions}

We have presented an elegant and general framework, called Posterior Matching, that allows VAEs to perform arbitrary conditioning. That is, we can take an existing VAE that only models the joint distribution $p(\x)$ and train an additional model that, when combined with the VAE, is able to assess any likelihood $p(\xu \mid \xo)$ for arbitrary subsets of unobserved features $\xu$ and observed features $\xo$.

We applied this approach to a variety of VAEs for a multitude of different tasks. We found that Posterior Matching outperforms previous specialized VAEs for arbitrary conditioning with tabular data and for image inpainting. Importantly, we find that one can switch to a more powerful base VAE and get immediate improvements in downstream arbitrary conditioning performance ``for free,'' without making changes to Posterior Matching itself. We can also use Posterior Matching to perform clustering based on partially observed inputs and to improve the efficiency of greedy active feature acquisition by several orders of magnitude at negligible cost to performance.

With this work, we hope to make arbitrary conditioning more widely accessible. Arbitrary conditioning no longer requires specialized methods, but can instead be achieved by applying one general framework to common VAEs. As advances are made in VAEs for joint density estimation, we can expect to immediately reap the rewards for arbitrary conditioning.

\begin{ack}
We would like to thank Google's TPU Research Cloud program for providing free access to TPUs. This research was partly funded by NSF grant IIS2133595 and by NIH 1R01AA02687901A1.
\end{ack}

\bibliography{references}

\newpage

\appendix

\section{Proof of \autoref{thm:pm-ac}}

\begin{proof}
Starting from our original objective
{\tiny
\begin{equation}
    \argmin_{\theta} \mathbb{E}_{\xu \sim p(\cdot \mid \xo)} \left[ \text{KL} \left( q_{\psi}(\z \mid \xo, \xu) \mid \mid q_{\theta}(\z \mid \xo) \right) \right],
\end{equation}
}%
we proceed as follows:
{\tiny
\begin{align}
    &= \argmin_{\theta} \mathbb{E}_{\xu \sim p(\cdot \mid \xo)} \left[ \int q_\psi(\z \mid \xo, \xu) \log \dfrac{q_\psi(\z \mid \xo, \xu)}{q_\theta(\z \mid \xo)} \dif \z \right] \\
    &= \argmin_{\theta} \mathbb{E}_{\xu \sim p(\cdot \mid \xo)} \left[ \int q_\psi(\z \mid \xo, \xu) \log \dfrac{q_\psi(\z \mid \xo, \xu)}{\dfrac{p_\phi(\xu \mid \z, \xo) q_\theta(\z \mid \xo)}{p_\phi(\xu \mid \z, \xo)}} \dif \z \right] \\
    &= \argmin_{\theta} \mathbb{E}_{\xu \sim p(\cdot \mid \xo)} \left[ \int q_\psi(\z \mid \xo, \xu) \log \dfrac{q_\psi(\z \mid \xo, \xu)}{\dfrac{p_{\theta,\phi}(\xu, \z \mid \xo)}{p_\phi(\xu \mid \z, \xo)}} \dif \z \right] \\
    &= \argmin_{\theta} \mathbb{E}_{\xu \sim p(\cdot \mid \xo)} \left[ \int q_\psi(\z \mid \xo, \xu) \log \dfrac{q_\psi(\z \mid \xo, \xu)}{\dfrac{p_{\theta,\phi}(\xu \mid \xo)q_\theta(\z \mid \xo, \xu)}{p_\phi(\xu \mid \z, \xo)}} \dif \z \right] \\
    &= \argmin_{\theta} \mathbb{E}_{\xu \sim p(\cdot \mid \xo)} \left[ \int q_\psi(\z \mid \xo, \xu) \log \dfrac{p_\phi(\xu \mid \z, \xo) q_\psi(\z \mid \xo, \xu)}{p_{\theta,\phi}(\xu \mid \xo)q_\theta(\z \mid \xo, \xu)} \dif \z \right] \\
    &= \argmin_{\theta} \mathbb{E}_{\xu \sim p(\cdot \mid \xo)} \left[ \int q_\psi(\z \mid \xo, \xu) \left( \log \dfrac{q_\psi(\z \mid \xo, \xu)}{q_\theta(\z \mid \xo, \xu)} + \log p_\phi(\xu \mid \z, \xo) - \log p_{\theta,\phi}(\xu \mid \xo) \right) \dif \z \right] \\
    &= \argmin_{\theta} \mathbb{E}_{\xu \sim p(\cdot \mid \xo)} \left[ - \log p_{\theta,\phi}(\xu \mid \xo) + \int q_\psi(\z \mid \xo, \xu) \left( \log \dfrac{q_\psi(\z \mid \xo, \xu)}{q_\theta(\z \mid \xo, \xu)} + \log p_\phi(\xu \mid \z, \xo) \right) \dif \z \right] \\
    &= \argmin_{\theta} \mathbb{E}_{\xu \sim p(\cdot \mid \xo)} \left[ - \log p_{\theta,\phi}(\xu \mid \xo) + \text{KL}\Big(q_\psi(\z \mid \xo, \xu) \mid \mid q_\theta(\z \mid \xo, \xu)\Big) + \mathbb{E}_{\z \sim q_\psi(\cdot \mid \xo, \xu)} \left[ \log p_\phi(\xu \mid \z, \xo) \right]\right] \\
    &= \argmin_{\theta} \mathbb{E}_{\xu \sim p(\cdot \mid \xo)} \left[ - \log p_{\theta,\phi}(\xu \mid \xo) + \text{KL}\Big(q_\psi(\z \mid \xo, \xu) \mid \mid q_\theta(\z \mid \xo, \xu)\Big) \right].
\end{align}
}%
In the final step, we can ignore the term $\mathbb{E}_{\z \sim q_\psi(\cdot \mid \xo, \xu)} \left[ \log p_\phi(\xu \mid \z, \xo) \right]$ since it does not depend on $\theta$. 
\end{proof}

\section{Likelihood Estimation} \label{sec:ll-est}

Often, it is of interest to evaluate likelihoods with generative models. In the VAE setting, the joint likelihood is frequently estimated with importance sampling:
\begin{equation}
    p_{\psi, \phi}(\x) = \mathbb{E}_{\z \sim q_\psi(\cdot \mid \x)} \Bigg[ \dfrac{p_\phi(\x \mid \z) p(\z)}{q_\psi(\z \mid \x)} \Bigg].
\end{equation}

With Posterior Matching, we can estimate arbitrary conditional likelihoods by additionally estimating
\begin{align}
    p_{\theta, \phi}(\xo) &= \mathbb{E}_{\z \sim q_\theta(\cdot \mid \xo)} \Bigg[ \dfrac{p_\phi(\xo \mid \z) p(\z)}{q_\theta(\z \mid \xo)} \Bigg] \\
    &= \mathbb{E}_{\z \sim q_\theta(\cdot \mid \xo)} \Bigg[ \dfrac{\int p_\phi(\x = (\xo, \xu) \mid \z) \dif \xu ~p(\z)}{q_\theta(\z \mid \xo)} \Bigg] \label{eq:xo-est}
\end{align}
in order to obtain $p_{\theta,\psi,\phi}(\xu \mid \xo) = p_{\psi, \phi}(\x)/p_{\theta, \phi}(\xo)$. This is the estimator we use in our experiments that report likelihoods. Note that if the decoder is factorized (which it always is in our experiments), then we can write \autoref{eq:xo-est} as:
\begin{equation}
    \mathbb{E}_{\z \sim q_\theta(\cdot \mid \xo)} \Bigg[ \dfrac{\prod_{i \in o}p_\phi(x_i \mid \z) p(\z)}{q_\theta(\z \mid \xo)} \Bigg].
\end{equation}
If the decoder is not factorized, then the integral in \autoref{eq:xo-est} needs to be estimated.

\section{Lookahead Posteriors}
~
\begin{figure}[h!]
    \centering
    \includegraphics[width=\linewidth]{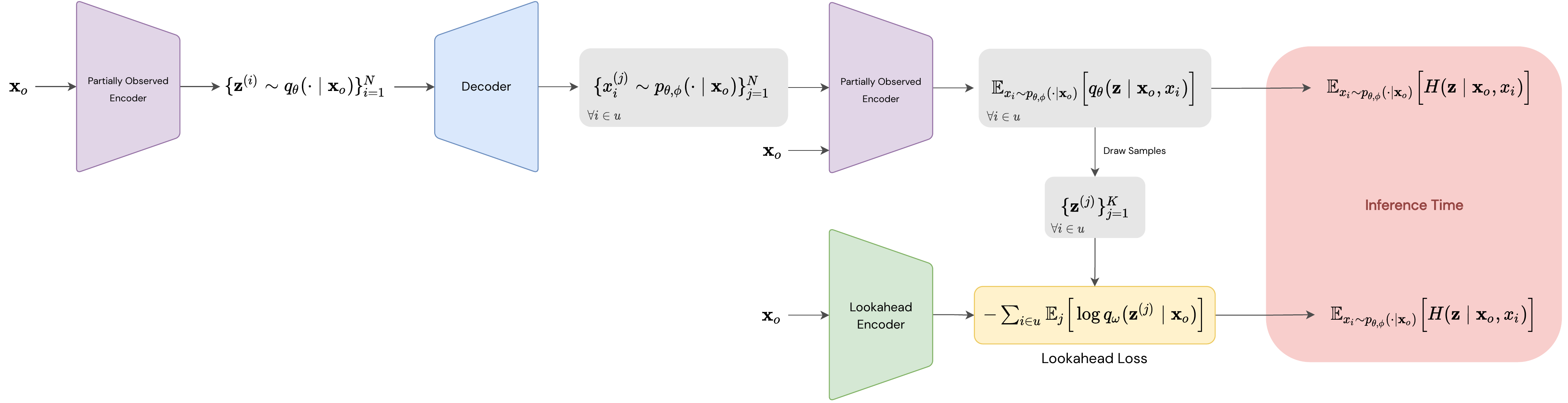}
    \caption{Overview of how Posterior Matching is used to learn ``lookahead'' posteriors for greedy active feature acquisition. The top path illustrates how samples are produced for use in the lookahead Posterior Matching loss. At inference time, we obtain the entropy of the distributions outputted by the rightmost networks. Taking the top path to obtain these entropies is the more expensive sampling-based approach. Using the learned Lookahead Encoder on the bottom path is the much faster approach that only requires a single network evaluation.}
    \label{fig:lookahead}
\end{figure}

\section{Experimental Details}

All experiments, except for VDVAE models (see \autoref{sec:hierarch-appendix}), were run on a single GeForce GTX 1080 Ti GPU, with the longest running models taking no more than 12 hours to train.

\subsection{Real-valued Datasets}

For experiments on the UCI datasets, we use multi-layer perceptrons (MLP) with residual connections for all networks. For these experiments, we found that only optimizing $\mathcal{L}_{\text{PM}}$ with respect to $\theta$ gave the best results (i.e.,~we stop gradients on samples from $q_\psi(\z \mid \xo)$ when computing $\mathcal{L}_{\text{PM}}$). We also use a schedule for the $\beta$ coefficient of the KL term in the ELBO, as this helped avoid degeneracy at the start of training. For models where a cyclical schedule \cite{fu2019cyclical} was used, the period was 50000 training steps and the schedule began after an initial 1000 steps where $\beta$ was 0 (except for \textsc{Miniboone}, where the period is 5000 and the delay is 2000). For models where a monotonic schedule was used, $\beta$ was 0 for the first 30000 training steps and then linearly annealed to 1 at the final training step. We used the autoregressive distribution described in \autoref{sec:auto-gmm}, with 256 hidden units and 3 residual blocks, for the partially observed posterior. During training, a small amount of Gaussian noise ($\sigma = 0.001$) is added to each minibatch. We use the Adam \cite{kingma2014adam} optimizer with an initial learning rate of 0.001 and an exponential decay schedule with a rate of 0.9 every 5000 steps (except for \textsc{Miniboone}, where the decay is every 1000 steps). During training and test time, observed masks were drawn from a Bernoulli distribution with $p=0.5$. At test time, likelihoods are computed with the estimator in \autoref{sec:ll-est}. Additional hyperparameters are given in \autoref{tab:uci-hparams}.

\begin{table}[h]
\centering
\caption{Additional hyperparameters for UCI experiments. Hidden Units, Residual Blocks, and Layer Normalization refer to the VAE encoder/decoder networks.}
\label{tab:uci-hparams}
\begin{tabular}{@{}lccccc@{}}
\toprule
                    & \textsc{Power}    & \textsc{Gas}      & \textsc{Hepmass}  & \textsc{Miniboone} & BSDS      \\ \midrule
Batch Size          & 512      & 512      & 512      & 1024      & 1024      \\
Latent Dimension    & 16       & 16       & 16       & 32        & 64        \\
Hidden Units        & 256      & 256      & 256      & 256       & 256       \\
Residual Blocks     & 2        & 2        & 2        & 5         & 5         \\
Layer Normalization & No       & No       & No       & Yes       & Yes       \\
Training Steps      & 200000   & 200000   & 200000   & 22000     & 200000    \\
$\beta$ Schedule    & Cyclical & Cyclical & Cyclical & Cyclical  & Monotonic \\ \bottomrule
\end{tabular}
\end{table}

\subsection{Image Inpainting}

\subsubsection{Vector Quantized-VAEs}

We first train VQ-VAE models as described in \citet{oord2017neural}. The VQ-VAE encoder and decoder are convolutional networks with residual blocks, following the implementation found at \url{https://github.com/deepmind/dm-haiku/blob/main/examples/vqvae_example.ipynb}. We use two residual blocks with 32 hidden units in the residual layers.  We use the exponential moving average version of the VQ-VAE training procedure in order to update the quantized vectors. We use a decay rate of 0.99 and a commitment cost of 0.25. The quantized vectors have a dimensionality of 64. Our decoder outputs a multivariate Gaussian with covarance matrix that is a scalar multiple of the identity matrix, where the scalar is a learnable parameter. All models are trained with the Adam \cite{kingma2014adam} optimizer with learning rate 0.0003. Additional hyperparemeters for each of the datasets can be found in \autoref{tab:vqvae-hparams}.

We then train a conditional PixelCNN \cite{oord2016conditional} for each pretrained VQ-VAE model, where the PixelCNN is modeling the partially observed posterior. First, we use the VQ-VAE encoder to obtain the discrete indices $\z$ that correspond to a given $\x$. We then use an encoder network with the same architecture as the VQ-VAE encoder to map $\xo$ to a 512-dimensional conditioning vector. This vector is then used as conditioning input to the PixelCNN when computing the log-likelihood of $\z$. This gives us $-\log q(\z \mid \xo)$, which is our usual Posterior Matching loss. We use a learnable embedding lookup as the first layer in the PixelCNN in order to map $\z$ to continuous values. The PixelCNN outputs categorical logits (with as many classes as there are discrete latent vectors). Convolutional layers in the PixelCNN use 128 filters. For CelebA, 12 residual blocks are used, and 8 are used for MNIST and Omniglot. Dropout is used with a rate of 0.5. We use the Adam optimizer with an initial learning rate of 0.0003 and decay of 0.999995 every step. The batch size is 32. Models are trained for 150000 steps. When training and evaluating the PixelCNN models, we randomly generate the masks of observed values according to the same distributions used by \citet{li2020acflow}. When evaluating the models, we compute PSNR by averaging over 10 decoded samples from the partially observed posterior.

\begin{table}[h]
    \centering
    \caption{Dataset-dependent hyperparameters for VQ-VAE.}
    \label{tab:vqvae-hparams}

    \begin{tabular}{@{}llll@{}}
    \toprule
                     & MNIST & \textsc{Omniglot} & \textsc{CelebA} \\ \midrule
    Batch Size       & 32    & 32       & 64     \\
    \# of Embeddings & 256   & 256      & 512    \\
    Hidden Units     & 32    & 32       & 128    \\
    Training Steps   & 60000 & 60000    & 100000 \\ \bottomrule
    \end{tabular}
    
\end{table}

\subsubsection{Hierarchical VAEs} \label{sec:hierarch-appendix}

We use the hierarchical VAE architecture proposed by \citet{child2020very} by closely following their original implementation at: \url{https://github.com/openai/vdvae}. However, we make the following modifications to incorporate the partially observed posterior. First, a second encoder network is added which accepts $\xo$ as input. We then also add an additional residual block in each top-down block in the decoder to output $q(\z_i \mid \z_{<i}, \xo)$. These new residual blocks are identical to the ones used to output the posteriors, except they accept the activations from the partially observed encoder instead of the fully observed encoder.

We use Gaussians for the partially observed posteriors, and so we directly compute the KL-divergence between the posterior and partially observed posterior at each level in the hierarchy.
In the hierarchical setting, the full Posterior Matching KL-divergence can be computed as
{\tiny
\begin{equation}
     \text{KL}\left(q_\psi(\z \mid \xo, \xu) \mid \mid q_\theta(\z \mid \xo)\right) = \sum_{i=1}^L \mathbb{E}_{\,\mathbf{z}_{<i} \sim q_\psi(\cdot \mid \mathbf{x}_o, \mathbf{x}_u) } \left[ \text{KL}\left(q_\psi(\z_i \mid \z_{<i}, \xo, \xu) \mid \mid q_\theta(\z_i \mid \z_{<i}, \xo)\right) \right],
\end{equation}
}%
and so during training we approximate this by simply summing the individual KL terms from all of the levels (this is analogous to how the KL term in the ELBO is computed for VDVAE). We stop gradients in the model such that the Posterior Matching loss is only computed with respect to the parameters of the partially observed encoder and the residual blocks that output the partially observed posteriors (i.e. only the new parameters that we introduced to the model).

We follow the training setup used by \citet{child2020very} as well. Our MNIST and \textsc{Omniglot} models have 20 levels in the hierarchy, and convolutions use 192 filters. The models were trained for 500000 steps on 8 TPU-v2 cores, which took about 3 days. Our \textsc{CelebA} model has 46 levels in the hierarchy, and convolutions use 384 filters. The model was trained for 1000000 steps on 8 TPU-v3 cores, which took about 5.5 days. TPUs were provided by Google's TPU Research Cloud program.

When evaluating likelihoods, we used the importance sampling estimator described in \autoref{sec:ll-est} with 10,000 samples (estimates had converged with this number of samples).

\subsection{Partially Observed Clustering}

We implemented and trained VaDE models as described in \citet{jiang2016variational}. However, we use convolutional encoders and decoders in our experiments instead of fully-connected networks. Otherwise, all hyperparameters are the same as in \citet{jiang2016variational}. In a straightforward adaptation of how VaDE typically predicts the cluster for $\x$, we predict the cluster based on $\xo$ with:
\begin{equation} \label{eq:cluster-pred}
    q(c \mid \xo) = \mathbb{E}_{\z \sim q(\cdot \mid \xo)} \Bigg[ \dfrac{p(\z \mid c) p(c)}{\sum_{c^\prime}p(\z \mid c^\prime) p(c^\prime)} \Bigg].
\end{equation}
We use 50 samples when estimating the expectation in \autoref{eq:cluster-pred}.

Training the partially observed posterior network is then done as usual. We use the same network architecture as the VaDE model's encoder for this network. We use the autoregressive distribution described in \autoref{sec:auto-gmm} for the partially observed posterior, with 256 hidden units and 2 residual blocks. Observed masks are sampled from a uniform distribution during training.

The supervised baseline is trained by first using the pretrained VaDE model to predict the class label for each instance (based on fully observed information). Those labels are then used as the ground truth to train a supervised model (that accepts partially observed inputs) with a standard cross-entropy loss. We use the same network architecture as the VaDE encoder and the partially observed posterior network for the supervised classifier. As before, observed masks are randomly generated during the training of this classifier.

\subsection{Very Fast Greedy Feature Acquisition}

In the feature acquisition experiments, we use relatively simple VAE models. For flattened MNIST, our encoder and decoder are MLPs with hidden layers of sizes 50, 100, and 200. We use this same architecture for EDDI as well. For the convolutional model, the encoder has four layers with the following \texttt{(hidden units, kernel, stride)}: \texttt{(32, 3, 1), (32, 3, 2), (64, 3, 2), (64, 1, 1)}. The decoder has the layers \texttt{(64, 8, 1), (64, 5, 2), (32, 5, 1), (32, 5, 1), (1, 3, 1)}. For both versions, the encoder outputs a diagonal Gaussian posterior, and the decoder outputs Bernoulli distributions. The latent dimension is 10 for all models.

For the partially observed posterior, we use a network with the same architecture as the fully observed encoder. Since we want to be able to compute the posterior entropy analytically when computing the information gains, we also use a Gaussian for the partially observed posterior. However, rather than letting it be diagonal, we parameterize the Gaussian with a lower triangular matrix $L$ such that the covariance matrix is $C = L L^\top$. We do not stop gradients on samples from the VAE encoder when computing the Posterior Matching loss.

We first train the VAEs with Posterior Matching, before learning the lookahead posteriors. During training, we uniformly generate masks that set between 0\% and 20\% of the features as observed. Our models are trained for 200000 steps with a batch size of 128. We use the Adam \cite{kingma2014adam} optimizer with an initial learning rate of 0.001 and an exponential decay schedule with a rate of 0.9 every 5000 steps.

After the VAE with Posterior Matching has been trained, we then freeze this model and train the lookahead posterior network. This network outputs one diagonal Gaussian for each feature. Given that the number of features can be relatively large and we are limited by the memory of the GPU, we randomly select a subset of these distributions to update at each training step. That is, we subsample the terms in the sum in \autoref{eq:lookahead-loss}. For the MLP model, we subsample 128 indices, and for the convolutional model we subsample 32 indices. We then estimate the expectation in \autoref{eq:lookahead-loss} over multiple samples from the already trained VAE model. We found using multiple samples to be important for getting good performance. For the MLP model, we use 64 samples, and for the convolutional model we use 16 samples. The MLP model is trained for 50000 steps with a batch size of 64 and the convolutional model is trained for 60000 steps with a batch size of 32.  We again use the Adam optimizer with an initial learning rate of 0.001 and an exponential decay schedule with a rate of 0.9 every 5000 steps.

\subsection{Autoregressive Posterior Details} \label{sec:auto-gmm}

As described in the main text, one of the advantages of Posterior Matching is the freedom to use highly expressive distributions for the partially observed posterior, as we do not require it to be reparameterizable. Here, we describe an autoregressive distribution that we use in some of our experiments. It is based on the proposal distributions used in \citet{strauss2021arbitrary} and \citet{nash2019autoregressive}, which were shown by \citet{strauss2021arbitrary} to outperform prior state-of-the-art arbitrary conditioning methods for likelihoods and imputation, despite being very simple.

The distribution consists of an MLP with residual connections that outputs a mixture of Gaussians for each covariate. This network accepts partially observed inputs as well as a conditioning vector as input. In our case, the conditioning vector is the output of the partially observed posterior encoder. We then compute $q(\z \mid \xo)$ in an autoregressive fashion as
\begin{equation*}
    q(\z \mid \xo) = \prod_{i=1}^D q(z_i \mid \xo, \z_{<i}),
\end{equation*}
where $D$ is the dimensionality of $\z$. Each $q(z_i \mid \xo, \z_{<i})$ term is obtained from a separate evaluation of the autoregressive distribution's network, where the partially observed inputs change to reflect the appropriate $\z_{<i}$. The conditioning vector that represents $\xo$ remains constant for all of these evaluations though. Note that when computing likelihoods, these evaluations can be done efficiently in parallel. Sampling, however, requires an $O(D)$ sequential procedure. However, we do not need to sample this distribution during training, and $D$ is generally small anyway.

We chose this particular distribution for its combination of simplicity and good performance. However, we did not experiment extensively with other types of autoregressive distribution for the partially observed posterior. As previously noted, though, there is a large degree of flexibility in this choice. 

\section{Zero Imputation with Base VAE}

\begin{figure}[h]
    \centering
    \includegraphics[width=0.2\linewidth]{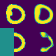}
    \includegraphics[width=0.2\linewidth]{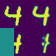}
    \includegraphics[width=0.2\linewidth]{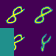}
    \caption{Demonstration of the zero-imputing approach to imputation with VAEs. In each image, the left column shows the inputs to a base VAE's encoder, and the right column shows the corresponding outputs. We show the masked regions of the inputs on the bottom row in green for clarity, even though those values are actually zero when passed to the encoder.}
    \label{fig:zero-imputation}
\end{figure}

One approach to imputation with VAEs that could be considered is to simply replace missing values with zeros before passing the input to the original VAE encoder and then subsequently using the decoder to obtain a reconstruction/imputation. While this method is straightforward and doesn't require any additional components in the model, it suffers greatly from distribution shift because the original VAE encoder was never trained to encounter those types of inputs. Thus, it is expected that this approach generally has worse performance. We include some simple results in \autoref{fig:zero-imputation} illustrating this, where we use the base VAE from our MNIST experiments. We see that the reconstructions of the zero-imputed inputs faithfully reproduce the zeros (i.e.~do not impute anything) and/or degrade the quality of the observed pixels' reconstructions.

\section{Additional Image Samples}

\begin{figure}[h]
    \centering
    \includegraphics{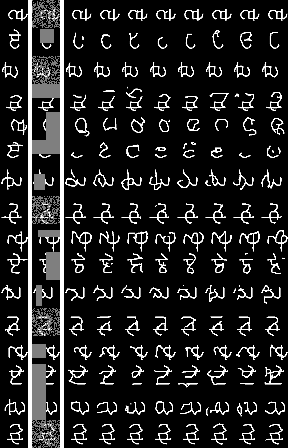}
    \caption{VDVAE \textsc{Omniglot} inpaintings.}
\end{figure}

\begin{figure}[p]
    \centering
    \includegraphics{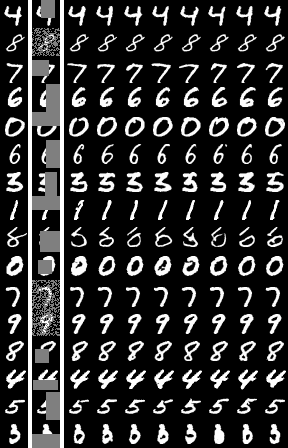}
    \caption{VDVAE \textsc{MNIST} inpaintings.}
\end{figure}

\begin{figure}[p]
    \centering
    \includegraphics[width=0.8\linewidth]{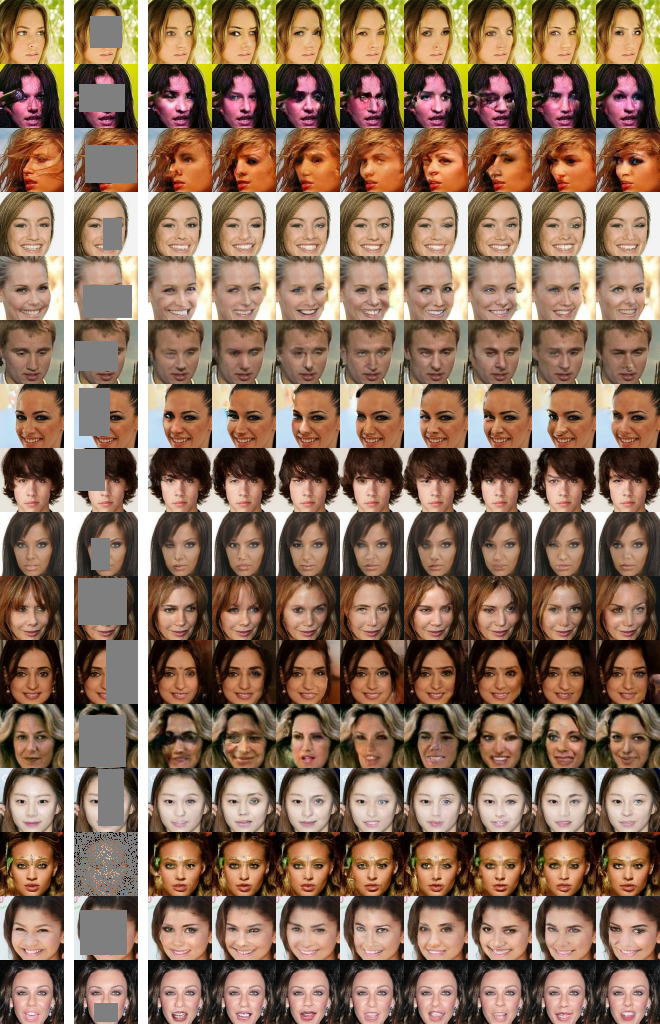}
    \caption{VDVAE \textsc{CelebA} inpaintings.}
\end{figure}

\begin{figure}[h]
    \centering
    \includegraphics{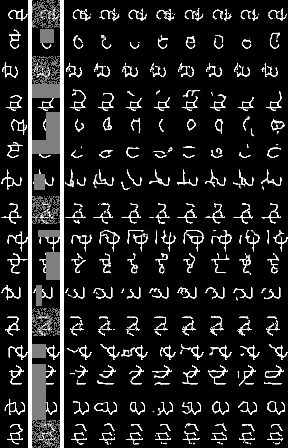}
    \caption{VQ-VAE \textsc{Omniglot} inpaintings.}
\end{figure}

\begin{figure}[p]
    \centering
    \includegraphics{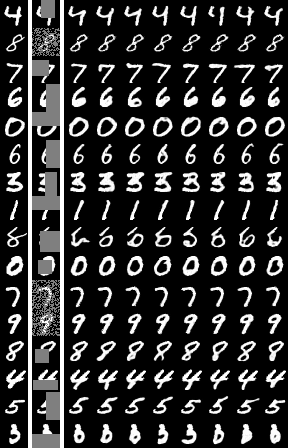}
    \caption{VQ-VAE \textsc{MNIST} inpaintings.}
\end{figure}

\begin{figure}[p]
    \centering
    \includegraphics[width=0.8\linewidth]{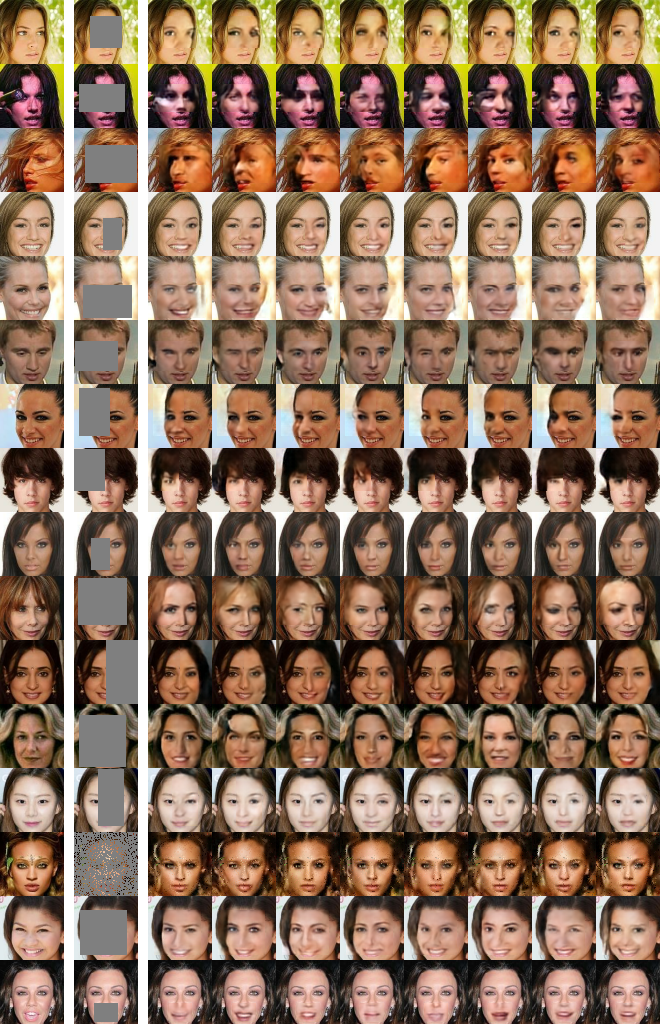}
    \caption{VQ-VAE \textsc{CelebA} inpaintings.}
\end{figure}

\begin{figure}[p]
    \centering
    \subfigure[\textsc{Omniglot}]{\includegraphics{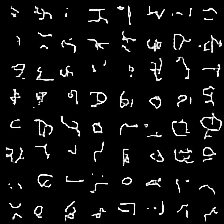}}
    \hskip 1cm
    \subfigure[\textsc{MNIST}]{\includegraphics{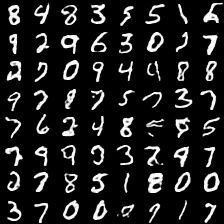}}
    \subfigure[\textsc{CelebA}]{\includegraphics{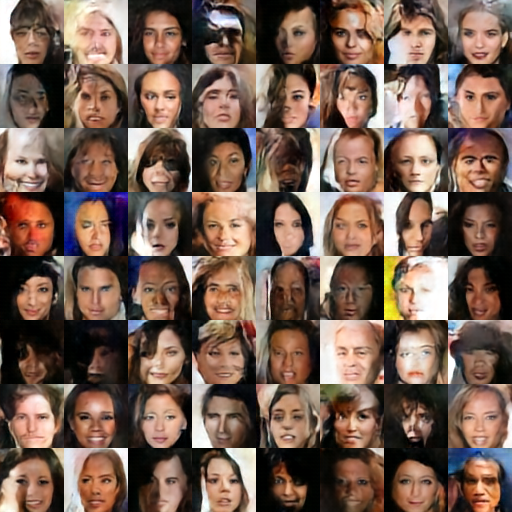}}
    \caption{VQ-VAE image samples from the joint distribution, a special case obtained by sampling $q_\theta(\z \mid \xo = \emptyset)$. Note that the models were not explicitly trained to model the joint and never saw $\xo = \emptyset$ during training.}
\end{figure}

\end{document}